\documentclass[10pt,twocolumn,letterpaper]{article}

\usepackage{cvpr}              

\usepackage{lipsum}
\usepackage{tikz}
\usetikzlibrary{backgrounds, fit, positioning}
\usepackage{adjustbox}
\usepackage{stmaryrd}
\usepackage{amsmath}
\usepackage{rotating}
\usepackage{pifont}
\usepackage{multirow}

\definecolor{myblueish}{RGB}{108, 142, 191}
\definecolor{myredish}{RGB}{184, 84, 80}

\definecolor{magenta}{HTML}{FF00FF}
\definecolor{cyan}{HTML}{00FFFF}
\newcommand{\cmark}{{\color{myblueish}\ding{51}}}%
\newcommand{\xmark}{{\color{myredish}\ding{55}}}%

\newcommand{\R}{\mathbb{R}}
\newcommand{\virg}[1]{``#1''}
\newcommand{\sset}[1]{\left\{ #1 \right\}} 
\newcommand{\stuple}[1]{\left( #1 \right)} 
\renewcommand{\vec}[1]{\boldsymbol{\mathrm{#1}}} 

\DeclareMathOperator*{\argmin}{arg\,min}

\usepackage{xspace}
\newcommand{\ourmethod}{EOGS\xspace}
\newcommand{\loc}{\text{loc}}
\newcommand{\homo}{\text{hom}}

\definecolor{cvprblue}{rgb}{0.21,0.49,0.74}
\usepackage[pagebackref,breaklinks,colorlinks,allcolors=cvprblue]{hyperref}

\def\paperID{10254} 
\def\confName{CVPR}
\def\confYear{2025}

\title{Gaussian Splatting for Efficient Satellite Image Photogrammetry}

\author{Luca Savant Aira$^1$ \quad\quad\quad\quad Gabriele Facciolo$^2$ \quad\quad\quad\quad Thibaud Ehret$^3$\\
{\tt\small \url{https://mezzelfo.github.io/EOGS/}}\\
\small $^1$ Politecnico di Torino, Corso Duca degli Abruzzi, 24, 10129 Torino TO, Italia\\
\small $^2$ Universite Paris-Saclay, CNRS, ENS Paris-Saclay, Centre Borelli, 91190, Gif-sur-Yvette, France\\
\small $^3$ AMIAD, Pôle Recherche, France
}

\begin{document}
\maketitle

\begin{abstract}
    Recently, Gaussian splatting has emerged as a strong alternative to NeRF, demonstrating impressive 3D modeling capabilities while requiring only a fraction of the training and rendering time. In this paper, we show how the standard Gaussian splatting framework can be adapted for remote sensing, retaining its high efficiency. This enables us to achieve state-of-the-art performance in just a few minutes, compared to the day-long optimization required by the best-performing NeRF-based Earth observation methods. The proposed framework incorporates remote-sensing improvements from EO-NeRF, such as radiometric correction and shadow modeling, while introducing novel components, including sparsity, view consistency, and opacity regularizations.
    \end{abstract}
    
    \begin{figure}[t]
     \begin{tabular}{c@{\hskip 0.1cm}c}
      \includegraphics[width=0.48\linewidth]{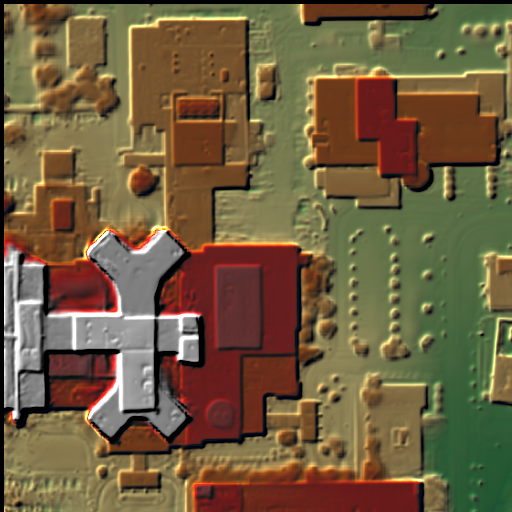} &
      \includegraphics[width=0.48\linewidth]{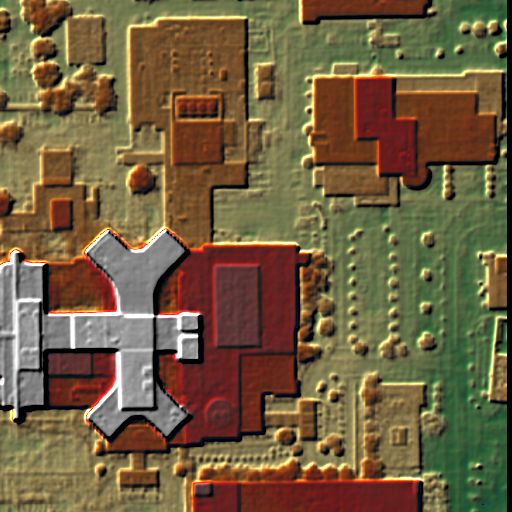} \\
      EOGS altitude & EO-NeRF altitude\\
      \includegraphics[width=0.48\linewidth]{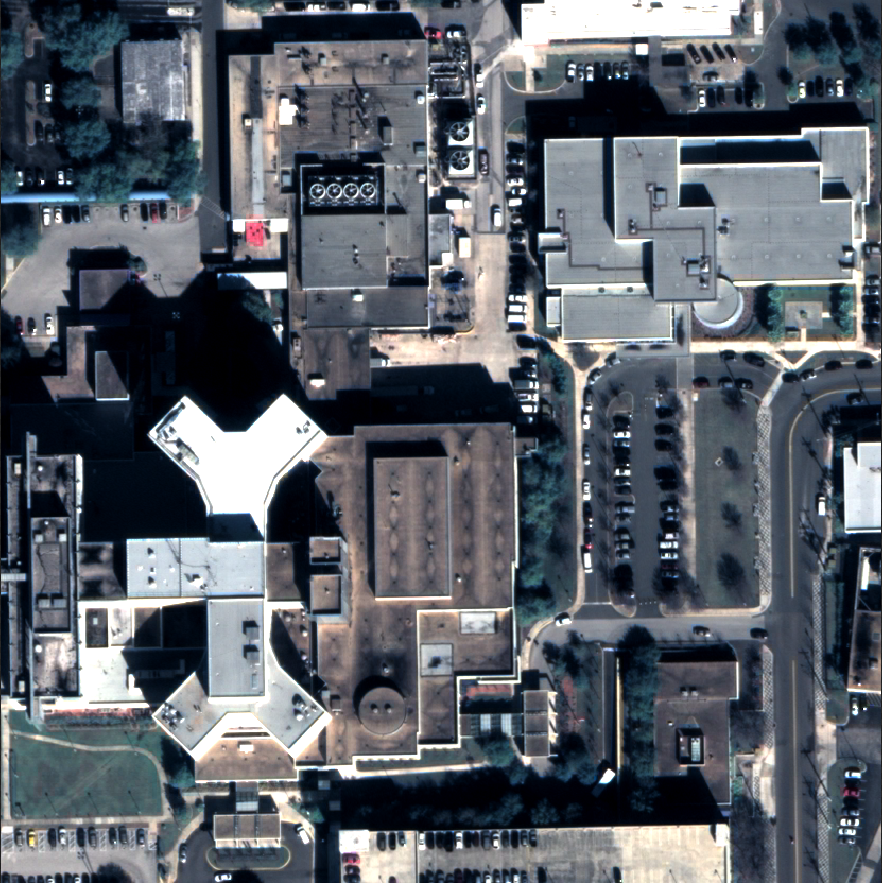} &
      \includegraphics[width=0.48\linewidth]{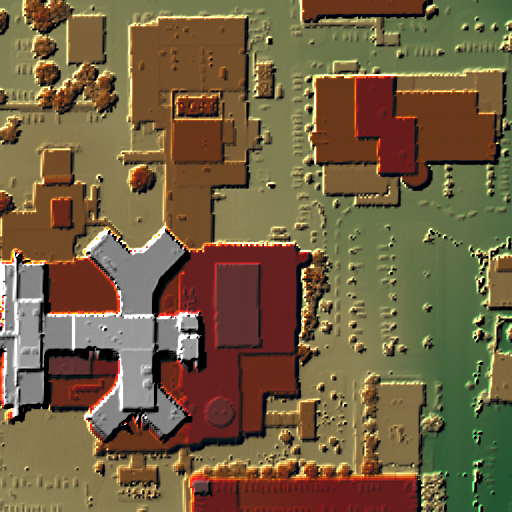} \\
      Pansharpened image & Ground-truth altitude\\
      \end{tabular}
      
    \vspace{-.5em}
      
      \caption{Using a limited number of satellite images of a given scene, the proposed \ourmethod method estimates the appearance and geometry of the scene. It achieves the same level of detail as EO-NeRF~\cite{mari2023eonerf}, such as the group of fans or the thin structures on top of the tall building on the left. However, EOGS requires only a few minutes of optimization, compared to the day-long training time required by EO-NeRF~\cite{mari2023eonerf}.}
      \label{fig:teaser}
    \end{figure}
    
    \section{Introduction}
    \label{sec:introduction}
    
    Since the mid-20th century, the number of active satellites in orbit has increased exponentially. Today, hundreds of satellites are dedicated to Earth observation, and this number is expected to continue growing~\cite{USGS_Compendium_2024}. These satellites frequently acquire optical images of the same areas at regular intervals. As a result, the availability of these datasets will continue to grow in the near future, requiring efficient algorithms to handle this expansion.
    
    One common task for these image datasets is photogrammetry~\cite{hartley1993relationship,lebegue2020co3d, WorldViewLegion_eoPortal}, which aims to recover the 3D geometry (\eg Digital Surface Model - DSM), and the appearance (\eg an albedo map) of the Earth's surface for this area using only the available 2D satellite images. Specifically, we focus on performing digital surface modeling from remote sensing images.
    
    Historically, binocular stereovision and tri-stereo methods have been used for this purpose. However, these methods rely on image acquisitions being nearly simultaneous and with specific relative positions, which is often impractical, with limited acquisition opportunities, and/or costly. Therefore, our goal is to develop a method capable of handling multi-date images captured from arbitrary satellite positions.
     
    More recently, multi-view stereo methods tailored for novel-view synthesis (NVS) have been used to solve this task as they naturally handle diverse camera positions. Indeed, as the NVS task also requires understanding the geometry of the scene, these methods can be used to recover a digital elevation model. These methods are usually based on the concept of radiance field, \ie a representation of the scene modeling the radiance emitted by each point, of which NeRF~\cite{mildenhall2021nerf} is the main representative. Among those, EO-NeRF~\cite{mari2023eonerf} has established itself as a state-of-the-art approach for digital surface modeling thanks to an improved shadow model. However, because it is NeRF-based, it also inherits NeRF’s computational slowness. In recent years, 3D Gaussian Splatting (3DGS)~\cite{kerbl3DGS} has been proposed as an alternative to NeRF. This method offers much faster training and rendering while reaching comparable reconstruction accuracy. 
    
    In this work, we introduce \ourmethod, the Earth-observation Gaussian splatting, the first method for digital elevation modeling based on 3DGS. \ourmethod achieves accuracy comparable to previous state-of-the-art approaches while being approximately $300\times$ faster. Keys to the success of \ourmethod are the following contributions, all of which are compatible with the original 3DGS framework's efficiency:
    \begin{itemize}
        \item Approximating locally the pushbroom satellite sensors as affine cameras.
        \item Introducing a shadow-mapping-based pipeline for rendering the shadows in a physically accurate manner.
        \item Adding three new regularization terms that promote sparsity in the Gaussians opacities, view consistency, and completely opaque objects. This reduces the training time and improves the quality of the results.
    \end{itemize}
    
    \section{Related Work}
    \label{sec:related_work}
    \subsection{Stereovision for Earth Observation}
    Stereovision is at the heart of many tools for 3D estimation from series of satellite images. Examples of such pipelines are Ames stereo pipeline~\cite{beyer2018ames}, MicMac~\cite{rupnik2017micmac}, CARS~\cite{michel2020new}, S2P~\cite{defranchis2014automatic}, or CATENA~\cite{krauss2013fully}.
    %
    %
    Traditionally, these multi-view stereo methods are applied to well-chosen (either manually or automatically) image pairs. Since they process each pair independently (for example in the dense stereo matching step), a crucial step is the fusion of all the generated pairwise 3D models into a single one.
    
    The recent trend has been to replace classic dense matching methods, such as semi-global matching (SGM)~\cite{hirschmuller2007stereo} or more global matching (MGM)~\cite{facciolo2015mgm}, with deep learning based methods such as PSM~\cite{chang2018pyramid}, HSM~\cite{yang2019hierarchical} or GA-Net~\cite{zhang2019ganet}. A review of these methods and a comparative study for satellite images is performed in \cite{mari2022disparity}.
    
    

    \subsection{NeRF for Earth Observation}
    
    Recently, Mildenhall \etal~\cite{mildenhall2021nerf} have shown that it is possible to learn a volumetric model of a scene, called neural radiance fields (NeRF), using differentiable inverse rendering. Given a sparse set of views of the scene, NeRF learns in a self-supervised manner by maximizing the photoconsistency across the predicted renderings corresponding to the available viewpoints. After convergence, the volumetric model can then be used to render realistic novel views of the scene. In practice, this volumetric model is represented by an MLP that predicts, for each position $\textbf{p}$ of the space, the local density of the scene $\sigma(\textbf{p})$ as well as its appearance (\ie color) $c(\textbf{p})$. The rendering is performed using an approximation of the volumetric integral $I$ from optical physics estimated using ray casting,
    \begin{equation}\label{eq:volumetric_integral}
        I(\vec{o}, \vec{d}) = \int_{t_n}^{t_f} e^{-\int_{t_n}^{t} \sigma\left(\vec{r}(s)\right) ds} c\left(\vec{r}(t), \vec{d} \right) \sigma\left(\vec{r}(t)\right) dt
    \end{equation}
    with $\vec{o}$ the camera center from which the ray $r$ originates and $\vec{d}$ unit direction.
    
    NeRF-based methods were then extended to the remote sensing case, and in particular to perform multi-view and multi-date satellite photogrammetry, namely S-NeRF~\cite{derksen2021shadownerf}, Sat-NeRF~\cite{mari2022satnerf}, and EO-NeRF~\cite{mari2023eonerf}.
    S-NeRF~\cite{derksen2021shadownerf} exploits the solar direction, information typically available in the metadata of each observation or that can be easily retrieved knowing the location of the scene as well as the acquisition hour and date, to predict the direct sun light reaching each point in the scene. This is done by adding the solar direction as an input to the MLP and predicting the amount of sun light reaching a point as a new output. In this way, the shadows cast by buildings can be learned by the MLP and generated accordingly during the novel-view rendering step.
    Sat-NeRF~\cite{mari2022satnerf} extends S-NeRF by modeling the transient parts of the scenes (\eg cars, construction sites, or foliage) as done in NeRF-in-the-wild~\cite{martin2021nerfW} and improves the camera representation of S-NeRF, from pinhole to RPC~\cite{tao2001comprehensive,baltsavias1992metric}.
    EO-NeRF~\cite{mari2023eonerf} improves the handling of shadows of S-NeRF and Sat-NeRF by defining physically plausible shadows directly from the geometry. These shadows are then rendered by additional raycasting from the surface in the direction of the sun.
    More recent work focuses on modeling difficult seasonal effects~\cite{gableman2024incorporating}, extending the proposed volumetric models to surface models~\cite{qu2023sat}, using the raw pre-pansharpened data provided directly by the satellite operators~\cite{pic2024pseudo}, and accelerating the training step by taking advantage of faster NeRF versions~\cite{billouard2024satngp}.

    \subsection{3D Gaussian Splatting}
    Following the growing interest in NVS, 3DGS~\cite{kerbl3DGS} was proposed as an alternative to NeRF-based methods. While both NeRF and 3DGS use alpha compositing as their image formation model, 3DGS represents the scene using a set of discrete Gaussian-shaped primitives placed in the 3D space, as opposed to a continuous black-box MLP representation of NeRF. In the following, we use the notation of Bulò~\etal~\cite{bulo2024revising}.
    
    A \textit{Gaussian primitive} is a tuple $\gamma = \stuple{\vec{\mu}, \Sigma, \alpha, \vec{f}}$ representing a single Gaussian-shaped volume element in the scene, where $\vec{\mu} \in \mathbb{R}^3$ is the primitive center, $\Sigma \in \mathcal{M}^{3\times 3}(\mathbb{R})$ its 3D shape and orientation, $\alpha \in [0,1]$ its opacity, and $\vec{f}\in \mathbb{R}^d$ its feature vector (\eg color when $d=3$).
    
    3DGS uses $K$ independent Gaussian primitives to represent the scene, so we will index them using $k=1,\ldots,K$ as subscript. Each Gaussian primitive $k$ has an associated 3D Gaussian kernel $\mathcal{G}_k$ defined as
    \begin{equation}
        \mathcal{G}_k (\vec{x}) = \exp{\left\{-\frac{1}{2} \left(\vec{x}-\vec{\mu}_k\right)^T \Sigma_k^{-1}(\vec{x}-\vec{\mu}_k) \right\}}.
    \end{equation}
    To render a view (characterized by its associated camera), 3DGS \virg{splats} each 3D Gaussian kernel onto the camera image plane. This process is called the \textit{splatting} operation and, mathematically, it associates a Gaussian primitive $\gamma_k$ and a camera model/projection $\mathcal{A} : \R^3 \to \R^2$ to a 2D Gaussian kernel $\mathcal{G}^{\mathcal{A}}_k : \R^2 \to \R$. The original 3DGS method deals only with pinhole camera models. During the splatting operation, each Gaussian is projected according to the first-order approximation of the perspective projection computed at $\vec{\mu}_k$, $J^{\mathcal{A}}\left(\vec{\mu}_k\right)$. In this way, the mean vector and covariance matrix of the 2D Gaussian kernel are:
    \begin{equation}\label{eq:definitionof2Dgaussiankernel3DGS}
        \vec{\mu}^{\mathcal{A}}_k = \mathcal{A}(\vec{\mu}_k) \qquad \Sigma^{\mathcal{A}}_k = J^{\mathcal{A}}\left(\vec{\mu}_k\right) \Sigma \left(J^{\mathcal{A}}\left(\vec{\mu}_k\right)\right)'.
    \end{equation}
    
    Once all the primitives are splatted, they are sorted front-to-back with respect to the camera reference. Then, they are aggregated using the traditional alpha compositing, accounting also for the Gaussian kernel decay such that
    \begin{equation}
        I^{\mathcal{A}} (\vec{u}) = \sum_{k=1}^{K} \Tilde{\vec{f}}_k \omega^{\mathcal{A}}_k (\vec{u}),
        \label{eq:3dgs_rendering}
    \end{equation}
    where $I^{\mathcal{A}}$ is the rendered image associated with camera $\mathcal{A}$, $\vec{u} \in \R^2$ is a point in the 2D image plane, $\Tilde{\vec{f}}_k$ are features that depend on $\vec{f}_k$ and the view direction (for modeling view-direction dependent color effect, such as shiny objects), and the alpha-compositing coefficient is
    \begin{equation}
        \omega^{\mathcal{A}}_k (\vec{u}) = \alpha_k \mathcal{G}^{\mathcal{A}}_k (\vec{u}) \prod_{j=1}^{k-1} \left(1-\alpha_j \mathcal{G}^{\mathcal{A}}_j (\vec{u})\right).
    \end{equation}
    
    Recent literature expanded the original 3DGS~\cite{kerbl3DGS} in many directions. As 3DGS has focused only on pinhole camera models and relied on a first-order local approximation of the model, recent works propose different camera models (\eg fish-eye~\cite{liao2024fisheye}) or more complex splatting strategies ~\cite{huang2024erroranalysis3dgaussian}. Another line of work focused on improving 3DGS in challenging input regimes, such as sparse-view input~\cite{xiong2023sparsegs}, and input without camera parameters~\cite{Fu_2024_CVPR}. Moreover, many recent methods focus on controllable texture and lighting~\cite{liang2023gs, R3DG2023, shi2023gir, jiang2024gaussianshader}. Further references can be found in recent surveys~\cite{wu2024recent, bao20243d}.
    
    
    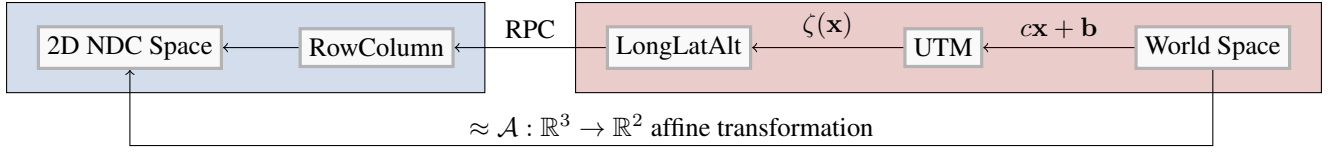
\begin{figure*}[ht]
        \centering
        \begin{tikzpicture}[
        squarenode/.style={rectangle, draw=gray!60, fill=gray!5, very thick, minimum size=5mm},
        align=center,
        ]
        \node[squarenode] (uvalt)                               {2D NDC Space};
        \node[squarenode] (rowcolalt)    [right=of uvalt]       {RowColumn};
        \node[squarenode] (longlatalt)   [right=2.0cm of rowcolalt]   {LongLatAlt};
        \node[squarenode] (UTM)          [right=2.0cm of longlatalt]  {UTM};
        \node[squarenode] (UTMnorm)      [right=2.0cm of UTM]         {World Space};
    
        \scoped[on background layer] \node (localcoordinates)  [draw, fill=myblueish!30, inner xsep=4mm, inner ysep=3mm, fit=(uvalt) (rowcolalt)] {};
        \scoped[on background layer] \node (worldcoordinates)  [draw, fill=myredish!30, inner xsep=4mm, inner ysep=3mm, fit=(longlatalt) (UTM) (UTMnorm)] {};
        
        \draw[->] (rowcolalt) -- (uvalt) node[midway,above] {}; 
        \draw[->] (longlatalt) -- (rowcolalt) node[midway,above] {RPC}; 
        \draw[->] (UTM) -- (longlatalt) node[midway,above] {$\zeta(\vec{x})$}; 
        \draw[->] (UTMnorm) -- (UTM) node[midway,above] {$c\vec{x}+\vec{b}$}; 
        \draw [->] (UTMnorm.south) |-([shift={(0mm,-10mm)}]UTMnorm.south) -- ([shift={(0mm,-10mm)}]uvalt.south) node[midway,above] {$\approx \mathcal{A}:\R^3\to\R^2$ affine transformation} -|(uvalt.south);
        \end{tikzpicture}
        \caption{Summary of the transformation from world-space to NDC-space and its affine approximation. The affine approximation is computationally efficient, compatible with the Gaussian splatting formulation, and well-suited for satellite images. The coordinate systems in the right red box represent 3D world coordinates (camera-independent), while the left blue box shows 2D coordinates (camera-dependent).}
        \label{fig:coordinatesystems}
    \end{figure*}
    
    \section{Method}
    \label{sec:method}
    
    The proposed Earth-observation Gaussian splatting method, referred to as \ourmethod, specializes and adapts 3DGS for the satellite photogrammetry task. Given $N$ non-orthorectified satellite images and their corresponding RPC camera model coefficients, a set of Gaussian-shaped 3D primitives is optimized to recover both the 3D geometry and appearance of the scene.
    
    The general learning problem is to find the set of $K$ Gaussian primitives that best approximates the $N$ satellite images, with the rendering process of \cref{eq:3dgs_rendering}. This can be formulated as:
    \begin{equation}\label{eq:generalproblem}
        \argmin_{(\gamma_k)_{\llbracket 1, K\rrbracket}} \sum_{i=1}^N \ell (\hat{I}_i, I_i),
    \end{equation}
    where $I_i$ is the $i$-th input satellite observation, $\hat{I}_i$ is the corresponding synthesized view (in the original 3DGS $\hat{I}_i = I^{\mathcal{A}_i}$), and $\ell$ is the same photometric distance function used in 3DGS.
    
    In the following sections we highlight the differences between \ourmethod and previous 3DGS and NeRF-based approaches.

    \subsection{Projections and Coordinate Systems}

    We define the coordinate system in which the Gaussian primitive centers and shapes are expressed as \textit{world-space coordinates}. This coordinate system is a uniformly rescaled and recentered version of the \textit{Universal Transverse Mercator (UTM) coordinate system}~\cite{utmref}, such that the center of the scenes coincides with the origin, the scene is contained in a unit cube, and it is east-north-up aligned similarly to EO-NeRF~\cite{mari2023eonerf}.
    At the other end of the transformation pipeline lies the \textit{2D NDC-space}, where the Gaussian primitives are splatted.
    
    The correct mapping between these two spaces is a composition of transformations: world-space to UTM to longitude-latitude-altitude. Using the RPC coefficients (that model the satellite position and 3D attitude) associated with each observation, the latter coordinate system is mapped onto the image row-column coordinates. Finally, the coordinates are normalized to range in $[-1,1]$ to get to NDC-space. As shown in \cref{fig:coordinatesystems}, we instead compute a per-scene affine approximation of the whole transformation, introducing a negligible mean error of $\approx 0.012$ pixels while being more computationally efficient than previous works and compatible with a Gaussian Splatting formulation. Specifically, for an affine camera model $\mathcal{A}:\vec{x}\in\R^3 \to A\vec{x}+\vec{a}\in\R^2$, the \cref{eq:definitionof2Dgaussiankernel3DGS} simplifies to
        $\vec{\mu}^{\mathcal{A}}_k = A\vec{\mu}_k+\vec{a} \in\R^2$  and
        $\Sigma^{\mathcal{A}}_k = A \Sigma A' \in \R^{2\times2}$.
    This simplification eliminates the need for the local first-order approximation used in the original 3DGS method, as we moved the approximation to the camera models.

    \subsection{Shadow Mapping}
    As in EO-NeRF, we want to explicitly model the shadow phenomena in the images, as the solar direction is available for each image in the scene. Unlike previous literature, our method uses a custom variant of shadow mapping to cast geometrically consistent shadows. Introduced in \cite{williams1978casting}, Shadow Mapping is a well-known technique in the field of 3D graphics for adding shadows to a computer graphic rendering. It is particularly suited for \ourmethod, as it requires just the ability to render the scene from different points of view, as opposed to the shadow casting technique used in EO-NeRF that requires ray marching (which is not defined in Gaussian splatting).
    
    Before introducing our variant of Shadow Mapping, we define the \textit{elevation render}, \textit{localization} function, and the \textit{homologous point} function.
    
    Given a camera model/projection $\mathcal{A}$, the elevation render is defined as the 3DGS rendering \cref{eq:3dgs_rendering} using the real elevations instead of colors 
    \begin{equation}
        E^{\mathcal{A}} (\vec{u}) = \sum_{k=1}^{K} \left[ \mathcal{E} \mu_k \right] \omega^{\mathcal{A}}_k (\vec{u}),
    \end{equation}
    where $\mathcal{E}:\R^3 \to \R$ is an affine operation mapping 3D points expressed in the \virg{native} world coordinates to the corresponding real altitude, expressed in meters. We remark that this is not the depth nor the inverse depth, typically found in the MVS literature.
    
    Given a camera model/projection $\mathcal{A}$, the localization function that maps a pixel of the camera and a given absolute altitude to its associated point in the native 3D world is defined as
    \begin{equation}
    \begin{aligned}
        \loc^\mathcal{A}: &\left(\R^2 \times \R\right) \to \R^3, \quad \left(\vec{u}, h\right) \mapsto \vec{x}
        \\\text{s.t.} \quad & \mathcal{A}(\vec{x}) = \vec{u} \quad \text{and} \quad \mathcal{E}(\vec{x}) = h.
    \end{aligned}
    \end{equation}
    
    Given two cameras $\mathcal{A}$ and $\mathcal{B}$, the homologous point function maps a pixel of the first camera to the corresponding pixel of the second camera, taking into consideration the 3D geometry:
    \begin{equation}
    \begin{aligned}
        \homo^{\mathcal{A},\mathcal{B}}:\ &\R^2 \to \R^2, \quad \vec{u} \mapsto \Tilde{\vec{u}}
        \\\text{s.t.} \quad & \Tilde{\vec{u}} = \mathcal{B}\left(\loc^\mathcal{A}\left(\vec{u}, E^{\mathcal{A}} (\vec{u})\right) \right).
    \end{aligned}
    \end{equation}
    
\begin{figure}[t]
        \centering
        \includegraphics[width=1\linewidth]{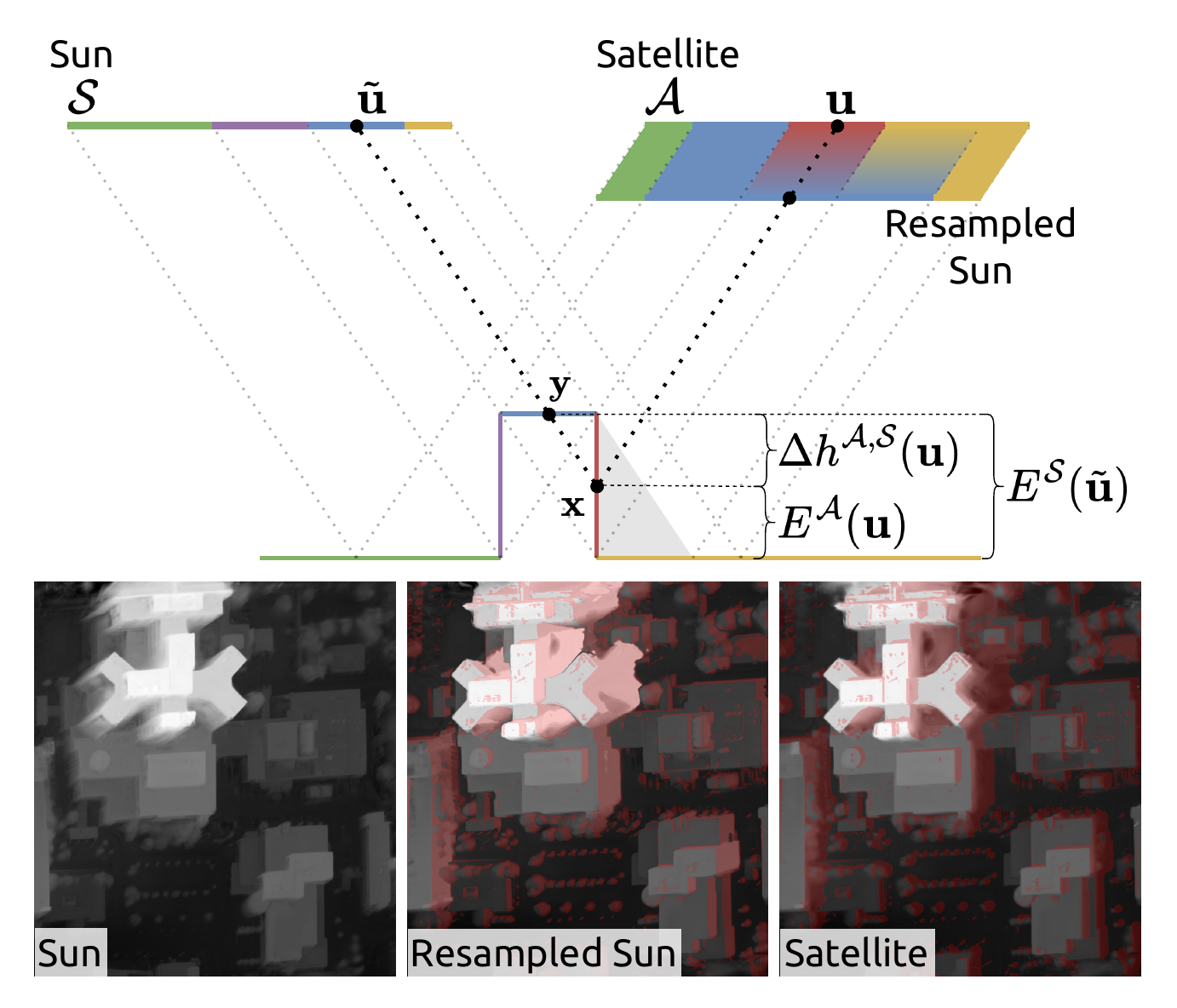}
    
        \vspace{-0.5em}
        
        \caption{Shadow mapping illustration. The point $\vec{u}$ in the satellite image (affine camera $\mathcal{A}$) corresponds to a the the 3D point $\vec{x}=\loc^\mathcal{A}(\vec{u})$ on the vertical wall. Projecting $\vec{x}$ to the sun camera (affine camera $\mathcal{S}$), $\Tilde{\vec{u}}=\mathcal{S}\vec{x}$ is obtained. Then $\vec{y}=\loc^\mathcal{S}(\Tilde{\vec{u}})$ is obtained localizing $\Tilde{\vec{u}}$. The point $\vec{x}$ and its pixel $\vec{u}$ are in shadow because the elevation of $\vec{y}$ is greater than the elevation of $\vec{x}$. Indeed, all and only the points where the satellite elevation and the resampled sun elevation do not match should be shaded.
        On the bottom of the illustration are shown examples of the sun elevation, the resampled sun elevation, and the satellite elevation renderings, with \textcolor{red}{shadows highlighted in red}.
        }
        \label{fig:shadow-mapping}
    \end{figure}

    In our shadow mapping approach (depicted in \cref{fig:shadow-mapping}), we assume that the sun is the only light source present in the scene. Moreover, since it is far from the scene, it can be approximated as a directional light. Following the classic shadow mapping approach, we construct a camera $\mathcal{S}$, called \textit{sun camera}, placed at and aligned with the light source (detailed in the supplementary). As the camera model corresponding to a directional light is the affine camera, we can handle uniformly the sun cameras and the satellite cameras.

    Then, we consider a second camera, $\mathcal{A}$, from which we want to synthesize a novel view and apply shadows according to the sun direction. Given a point $\vec{u}$ in the $\mathcal{A}$ NDC-space, and its corresponding altitude $E^{\mathcal{A}}(\vec{u})$, we first localize it, obtaining a 3D point in world-space. We then project this point according to $\mathcal{S}$, obtaining the homologous point of $\vec{u}$ in $\mathcal{S}$. We then resample the elevation rendering of $\mathcal{S}$ at this projected point and compare it with $E^{\mathcal{A}}(\vec{u})$. Mathematically, this corresponds to
    \begin{equation}
        \Delta h^{\mathcal{A},\mathcal{S}} (\vec{u})= E^{\mathcal{S}}
        \left(\homo^{\mathcal{A},\mathcal{S}}(\vec{u})\right) - E^{\mathcal{A}}(\vec{u}).
    \end{equation}
    If these two elevations, $E^{\mathcal{A}}(\vec{u})$ and $E^{\mathcal{S}}(\Tilde{\vec{u}})$ are the same, it means that both the camera and the sun camera are imaging the same 3D point, hence this point is in light. If the two elevations are not the same, then the sun camera is not able to \virg{see} the 3D point, hence it is in shadows. To represent this shading, the color of points in the shadows is multiplied by a darkening coefficient computed from $\Delta h^{\mathcal{A},\mathcal{S}} (\vec{u})$ as
    \begin{equation}
    s^{\mathcal{A},\mathcal{S}} (\vec{u}) = \min\sset{\exp{\left\{- \rho \Delta h^{\mathcal{A},\mathcal{S}} (\vec{u})\right\}},1}.
    \label{eq:darkening_from_Deltah}
    \end{equation}
    We argue that this formulation is physically plausible as this would be the correct equation for a homogeneous medium of density $\rho$, as shown in \cite{468400}.
    
    Following~\cite{mari2023eonerf}, we also model a per-camera ambient light $\vec{\psi}^{\mathcal{A}}$ so that in-shadow objects do not appear completely black. The shading to be applied to a given pixel $\vec{u}$ is given by the following \textit{lighting coefficient}
    \begin{equation}
        l^{\mathcal{A},\mathcal{S}} (\vec{u}) = s^{\mathcal{A},\mathcal{S}}(\vec{u})+(1-s^{\mathcal{A},\mathcal{S}}(\vec{u})) \vec{\psi}^{\mathcal{A}}.
    \end{equation}
    Finally, \ourmethod image formation equation is:
    \begin{equation}
        I^{\mathcal{A},\mathcal{S}} (\vec{u}) = l^{\mathcal{A},\mathcal{S}} (\vec{u}) \sum_{k=1}^{K} \vec{\phi}^{\mathcal{A}}(\vec{f}_k) \omega^{\mathcal{A}}_k (\vec{u}),
        \label{eq:im_form}
    \end{equation}
    where $\vec{\phi}^{\mathcal{A}}(\cdot)$ is a camera-specific affine color correction applied to the intrinsic primitive colors $\vec{f}_k$. We remark that, differently from 3DGS, we drop the view-direction dependencies of the primitive colors and introduce a camera-dependent color correction.
    
    It is useful to define the \textit{albedo rendering}, where we do not use the shadows or the camera-specific color correction:
    \begin{equation}
        I^{\mathcal{A}} (\vec{u}) = \sum_{k=1}^{K} \vec{f}_k \omega^{\mathcal{A}}_k (\vec{u}).
    \end{equation}
    While the image formation model defined in \cref{eq:im_form} is equivalent to EO-NeRF, the shadow definition is quite different.
    In EO-NeRF case, shadows are defined as the sun visibility for all points on the surface. Because of possible occlusions, two points of the scene can correspond to the same point seen from the sun direction. Therefore, it is not possible to define the sun visibility as an ``image" that could be estimated using a Gaussian splatting-like process. Trying to compute an irregularly sampled ``image" corresponding to these points would break the locality assumption used in Gaussian splatting during the rasterization step and thus reduce the computational efficiency.
    On the contrary, the proposed shadow mapping verifies all the assumptions made by Gaussian splatting.

    \subsection{Regularizers}
    It is well-known that deep neural networks are implicitly regularized \cite{ramasinghe2022frequency, poggio2018theory, tancik2020fourfeat, ulyanov2018deep}, meaning that despite being used in the overparametrized regime, they show generalization capabilities.
    
    On the other hand, we found out that 
    {primitives in 3DGS-based methods are almost independently optimized one from the other. This is probably due to the fact that the primitives in 3DGS are initialized as small spheres, spread out in the entire scene.}  This results in 3DGS being less regularized than NeRF-based methods and \virg{lacking} constraints during the optimization phase.
    
    Hence we are free to add additional regularization constraints to the general optimization problem \cref{eq:generalproblem} that induce smoother and more regular solutions. In particular, we introduce constraints that promote our solution to be sparse (\ie we encourage solutions that require fewer Gaussian primitives), view consistent, and mostly composed of completely opaque objects.
    
    As common ML pipelines are specialized for unconstrained optimization problems, we argue to use a Lagrangian relaxation approach and re-formulate each constraint as a new loss term, each with its own experimentally-found Lagrangian multiplier.

    \smallskip\noindent\textbf{Promoting Sparsity.}
    Training time is directly proportional to the number of Gaussian primitives considered during the optimization process. As we want to recover the geometry of the scene as fast as possible, we want as few Gaussian primitives as possible, hence a sparse solution.
    
    Inspired by the well-known LASSO regularization in linear regression~\cite{tibshirani1996regression} that promotes a sparse solution, and by recent works such as 3DGSMCMC~\cite{kheradmand20243d}, we consider a $L^1$ regularization of the opacities
    \begin{equation}
        \mathcal{L}_{o} = \frac{1}{K}\sum_{k=1}^{K} \alpha_k.
    \end{equation}
    
    This regularization promotes sparsity in the primitive opacities distribution, hence only \virg{useful} primitives will be visible at the end of the optimization. We pair this regularization with a simple thresholding pruning procedure that discards any primitive with $\alpha < \alpha_{\min}$. In this way, unused primitives are actually discarded, yielding faster splatting and overall faster training (specifically, we recorded speedups of up to $2\times$ on the considered datasets). 
    
    We remark that many works~\cite{bulo2024revising, kheradmand20243d} have proposed replacements to the original 3DGS densification/pruning procedure. Here, instead, we aim only at lowering the number of primitives, so we do not need a densification strategy as long as we instantiate enough of them at the beginning of the optimization. Moreover, we set $\alpha_{\min} = 0.0025$ as primitives with lower opacities are already discarded in the original 3DGS implementation of the front-to-back splatting procedure.

    \begin{figure}[t]
    \centering
      \includegraphics[width=0.38\linewidth]{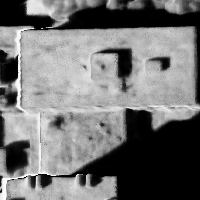}
      \includegraphics[width=0.38\linewidth]{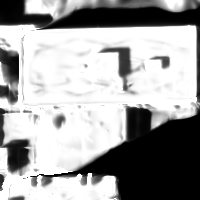}\\
      \includegraphics[width=0.38\linewidth]{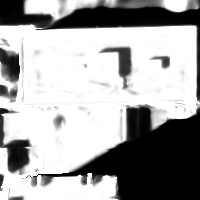}
      \includegraphics[width=0.38\linewidth]{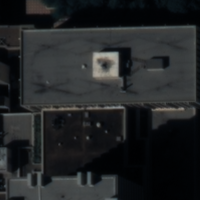}
      \caption{From top-left to bottom-right, shadow maps of EO-NeRF, EOGS without the $\mathcal{L}_s$ penalizer, EOGS with the $\mathcal{L}_s$ penalizer, and the corresponding satellite image. Textures corresponding to the image content can be observed in the shadow map of EO-NeRF and EOGS without the $\mathcal{L}_s$ penalizer, but not in EOGS.}
      \label{fig:shadow_penalizer}
      
      \vspace{-0.5cm}
      
    \end{figure}
    \smallskip\noindent\textbf{Promoting View Consistency.}
    Differently from the classical NVS context, in remote sensing the available views are low-count and sparse, resulting in \cref{eq:generalproblem} being even less constrained. Paired with the fact that 3DGS does not benefit from the implicit regularization of NeRF, we argue that an additional constraint promoting view consistency is needed. 
    
    We propose a \virg{local view consistency} loss based on the intuition that if the same 3D point is visible from two cameras and the cameras are close to each other, then the color and elevation resampled at the corresponding pixels should be the same. Otherwise, the object is occluded or outside the camera boundaries.
    
    
    Mathematically, given a camera $\mathcal{A}$ we randomly perturb it and obtain a camera $\mathcal{B}$. Assuming that there is no view-direction dependent color effect, this constraint reads:
    \begin{equation}
        \Delta h^{\mathcal{A},\mathcal{B}}(\vec{u}) < \Delta h_{\min} \Rightarrow \begin{cases}
            I^{\mathcal{A}} (\vec{u})
            =
            I^{\mathcal{B}}\left(\homo^{\mathcal{A},\mathcal{B}}(\vec{u})\right) \\
            E^{\mathcal{A}} (\vec{u})
            =
            E^{\mathcal{B}}\left(\homo^{\mathcal{A},\mathcal{B}}(\vec{u})\right),
        \end{cases}
    \end{equation}
    where we reused the same notation of the shadow mapping explanation.
    
    This constraint results in two loss terms, the color (albedo) consistency and the altitude consistency:
    \begin{equation}
        \mathcal{L}_{cc} = \sum_{\vec{u}} M^{\mathcal{A},\mathcal{B}}(\vec{u}) \left|I^{\mathcal{A}} (\vec{u}) - I^{\mathcal{B}}\left(\homo^{\mathcal{A},\mathcal{B}}(\vec{u})\right)\right|
    \end{equation}
    \begin{equation}
        \mathcal{L}_{ac} = \sum_{\vec{u}} M^{\mathcal{A},\mathcal{B}}(\vec{u}) \left|E^{\mathcal{A}} (\vec{u}) - E^{\mathcal{B}}\left(\homo^{\mathcal{A},\mathcal{B}}(\vec{u})\right)\right|,
    \end{equation}
    where $M^{\mathcal{A},\mathcal{B}}(\cdot)$ is a binary mask that selects all pixels $\vec{u}$ such that $\Delta h^{\mathcal{A},\mathcal{B}}(\vec{u}) < \Delta h_{\min}$ and $\homo^{\mathcal{A},\mathcal{B}}(\vec{u})$ is inside the image boundaries. 
    We remark that we always choose $\mathcal{A}$ from the input posed images and we set $\Delta h_{\min} = 30\text{cm}$. Moreover, we obtain $\mathcal{B}$ by independently sampling $q_1, q_2 \in \R$ from a $\pm1$-truncated standard distribution and defining
    \begin{equation}
        \mathcal{B}(\vec{x}) = \mathcal{A}(\vec{x})+0.05\ \mathcal{E}(\vec{x}) \begin{pmatrix}
            q_1 \\ q_2
        \end{pmatrix}.
    \end{equation}
    
    In the generic NVS literature, many works~\cite{niemeyer2022regnerf, darmon2022improving} proposed different methods for increasing the view consistency. RegNeRF~\cite{niemeyer2022regnerf} is the first work that deals with sparse camera poses by introducing a loss term that maximizes the likelihood of rendered RGB patches from virtual cameras with a pre-trained deep normalizing-flow model, while also adding a total variation regularization on the rendered depth. Furthermore, \cite{darmon2022improving} introduces a reprojection mechanism such that only the geometry needs to be learned from the NeRF, as the colors are resampled from the input images. Note that \ourmethod differs from \cite{niemeyer2022regnerf} as we ask for consistency (RGB and depth) between two cameras (one real and one virtual), hence we do not need any pre-trained model for the RGB renders nor prior on the elevation renders. \ourmethod also differs from \cite{darmon2022improving} as we learn the colors and do not resample input images that may contain transients or color shifts.
    

    \begin{table*}[ht]
    \centering
    \begin{tabular}{@{}lrcccccccccc@{}}
    & & \multicolumn{5}{c}{JAX} & \multicolumn{4}{c}{IARPA}     \\ \cline{3-7} \cline{8-11} 
    & & 004 & 068 & 214 & 260 & Mean $\downarrow$ & 001 & 002 & 003 & Mean $\downarrow$ & Time $\downarrow$ \\
    & Number of views     & 8    & 16   & 20  & 14  & - & 24      & 20     & 21  & - & -  \\
        \midrule
        \multirow{4}{*}{\rotatebox[origin=c]{90}{No mask}}
    & EO-NeRF~\cite{mari2023eonerf}           & 1.37 & 1.05 & 1.61 & 1.37 & 1.35 & 1.43 & 1.79 & 1.31  & 1.51 & 15 hours \\
    & SAT-NGP~\cite{billouard2024satngp}      & 1.63 & 1.27 & 2.18 & 1.79 & 1.72 & 1.54 & 2.11 & 1.69  & 1.78 & 25 minutes \\
    & Sat-Mesh~\cite{qu2023sat}               & 1.55 & 1.15 & 2.02 & 1.36 & 1.52 & N.A. & N.A. & N.A.  & N.A. & 8 minutes \\
    & S2P~\cite{facciolo2017automatic} & 1.45 & 1.19 & 1.82 & 1.66 & 1.53 & 1.48 & 2.48 & 1.38 & 1.78 & 20 minutes \\
    & \ourmethod (ours)                       & 1.45 & 1.10 & 1.73 & 1.55 & 1.46 & 1.58 & 2.00 & 1.27  & 1.62 & 3 minutes \\
        \midrule
        \multirow{3}{*}{\rotatebox[origin=c]{90}{\shortstack{Foliage\\mask}}}
    & EO-NeRF~\cite{mari2023eonerf}       & 1.02 & 1.03 & 1.55 & 1.24 & 1.21 & 1.32 & 1.63 & 1.18 & 1.38 & 15 hours   \\
    & SAT-NGP~\cite{billouard2024satngp}  & 1.03 & 1.26 & 2.17 & 1.43 & 1.47 & 1.34 & 1.85 & 1.62 & 1.60 & 25 minutes \\
    & \ourmethod (ours)                   & 0.89 & 1.01 & 1.63 & 1.24 & 1.19 & 1.38 & 1.70 & 1.03 & 1.37 & 3 minutes  \\
    \bottomrule
    \end{tabular}
        
        \vspace{-0.5em}
    
    \caption{Mean absolute error on the elevation [meters] and the corresponding training time for various baseline methods, when considering the whole AOI (no mask) or when ignoring foliage areas (foliage mask). Results for Sat-Mesh are reported from the paper since the authors did not share their code.}
    \label{tab:exp_main} 
    \end{table*}

    \smallskip\noindent\textbf{Promoting Opaqueness.}
    Looking at the output from SatNeRF and EO-NeRF (see \cref{fig:shadow_penalizer}), we can see that much of the texture of the scene is embedded in the geometric shadows. This geometry misuse is caused by semi-transparent objects casting semi-transparent shadows. In order to lessen this effect, we propose to add an entropy-based penalty $\mathcal{L}_s$ for incorrect use of the shadows. This penalty is defined as 
    \begin{equation}
        \mathcal{L}_s = \sum_{\vec{u}} \textit{H}\left(s^{\mathcal{A},\mathcal{S}} (\vec{u})\right),
    \end{equation}
    where $\textit{H}(x) = -\left(x \log_2(x)+(1-x)\log_2(1-x)\right)$.
    Indeed, the shadow map $s^{\mathcal{A},\mathcal{S}}$ should contain only $0$ or $1$ values. This is the case for $\rho \to +\infty$ in \cref{eq:darkening_from_Deltah}, as a building should not cast a semi-transparent shadow. Hence we add this entropy-based penalizer to discourage the use of semi-transparent shadows, which in turn encourage objects to be either completely transparent or fully opaque. Note that choosing a large $\rho$ during training is not an option since it would make the training unstable as \cref{eq:darkening_from_Deltah} would be close to a non-differentiable step function.
    
    We acknowledge that the problem of promoting hard surfaces in NeRF has been studied in previous works, such as~\cite{rebain2022lolnerf, barron2022mip}.

    \subsection{Implementation Details}
    The implementation of \ourmethod is based on the original 3DGS code base. Other than the aforementioned novel contributions, the main differences lie in disabling the per-Gaussian view-direction color dependency and initializing all the Gaussians with white color and as low as possible opacity ($1\%$). Moreover, we reduce the number of iterations to 5000 and enable the shadow mapping and all three regularizations at iteration 1000. Furthermore, the Gaussians centers are initialized uniformly in the 3D scene such that the initial density is $0.13$ Gaussians per $m^3$.
    
    We use the same optimizer and scheduler of 3DGS for the primitives and use a second Adam scheduler with $10^{-2}$ learning rate for learning the camera-dependent parameters: the affine color-correction $\vec{\phi}^{\mathcal{A}}$ and the ambient color $\vec{\psi}^{\mathcal{A}}$.
    
    The Lagrangian coefficients of the regularization constraints have been found experimentally on a single scene, rounded to the nearest power of ten, and applied to all scenes. This highlights the robustness of \ourmethod to the specific values of these coefficients. The final loss is:
    \begin{equation}\label{eq:regularizedproblem}
        \min \sum_{i=1}^N \ell (\hat{I}_i, I_i) + 0.1 \mathcal{L}_o + 0.1 \mathcal{L}_{cc} + 0.01 \mathcal{L}_{ac} + 0.01 \mathcal{L}_s,
    \end{equation}
    where $\hat{I}_i$ is now, differently from 3DGS in~\cref{eq:generalproblem}, a shorthand notation for $I^{\mathcal{A}_i,\mathcal{S}_i}$ from \cref{eq:im_form}, which also depends on the sun camera $\mathcal{S}_i$.
    
    
    \section{Experiments}
    \label{sec:experiments}

    We evaluate \ourmethod in the same experimental setting as the most recent related work in the literature, EO-NeRF.
    
    We are using datasets provided in the 2019 IEEE GRSS Data Fusion Contest (DFC2019)~\cite{bosch2019semantic,lesaux2019data} and 2016 IARPA Multi-View Stereo 3D Mapping Challenge (IARPA2016). These datasets, comprising a total of 7 areas of interest (AOI), contain cropped non-orthorectified multidate WorldView-3  observations, along with metadata such as the 3D satellite attitude (encoded in the RPC coefficient) and the local sun direction. We use the bundled-adjusted version of the RPC coefficient used in EO-NeRF. Each image covers approximately $256\times256$ meters squared of terrain with a resolution of $30 \sim 50$ cm per pixel, while each AOI is imagined by $10 \sim 20$ crops.
    
    \subsection{Main Experiment Results}
    \cref{tab:exp_main} show the main experimental results of \ourmethod. To assess the accuracy of \ourmethod we report the mean absolute error (MAE) between a lidar scan included in the dataset and the elevation render aligned to this nadir view. We argue that the volume of these data will grow in the near future, so we are also interested in the time required to recover the geometry from the input images. Hence, we also report the training time. 
    
\begin{figure}
        \centering
        \includegraphics[width=0.24\linewidth]{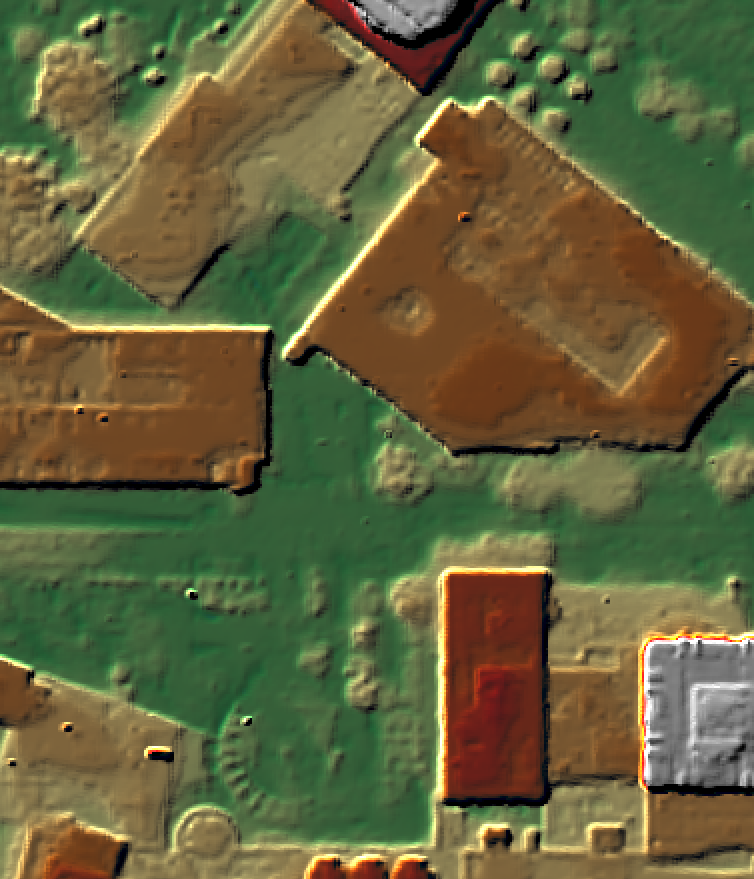}
        \includegraphics[width=0.24\linewidth]{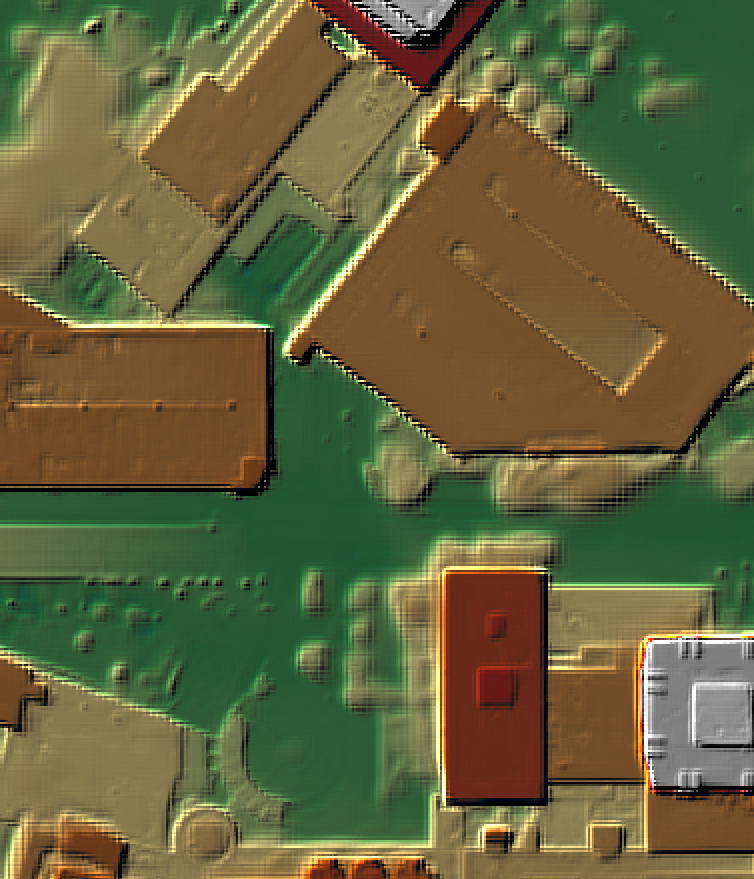}
        \includegraphics[width=0.24\linewidth]{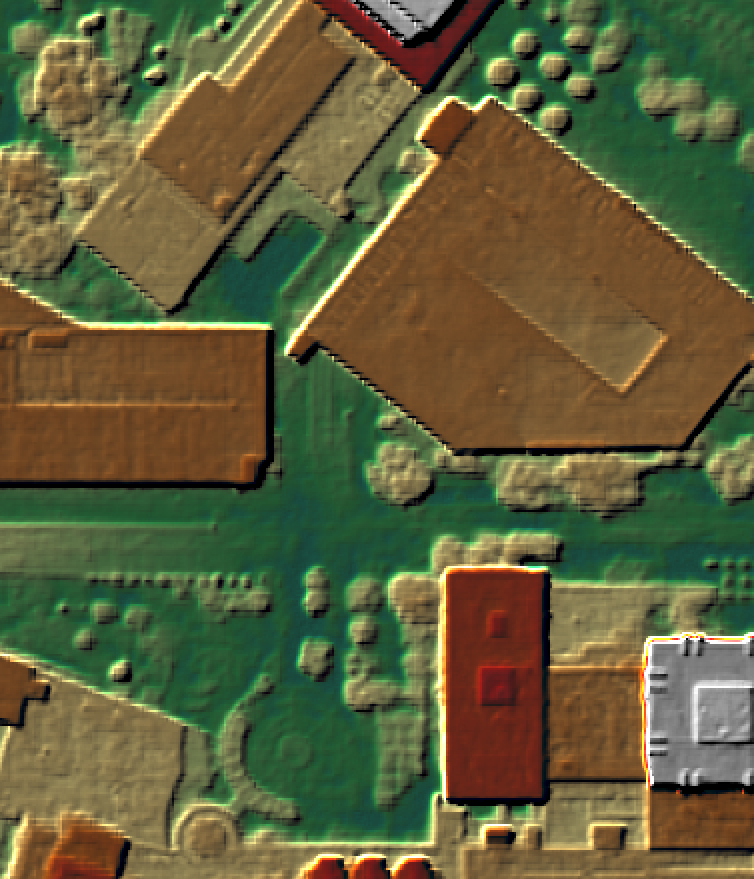}
        \includegraphics[width=0.24\linewidth]{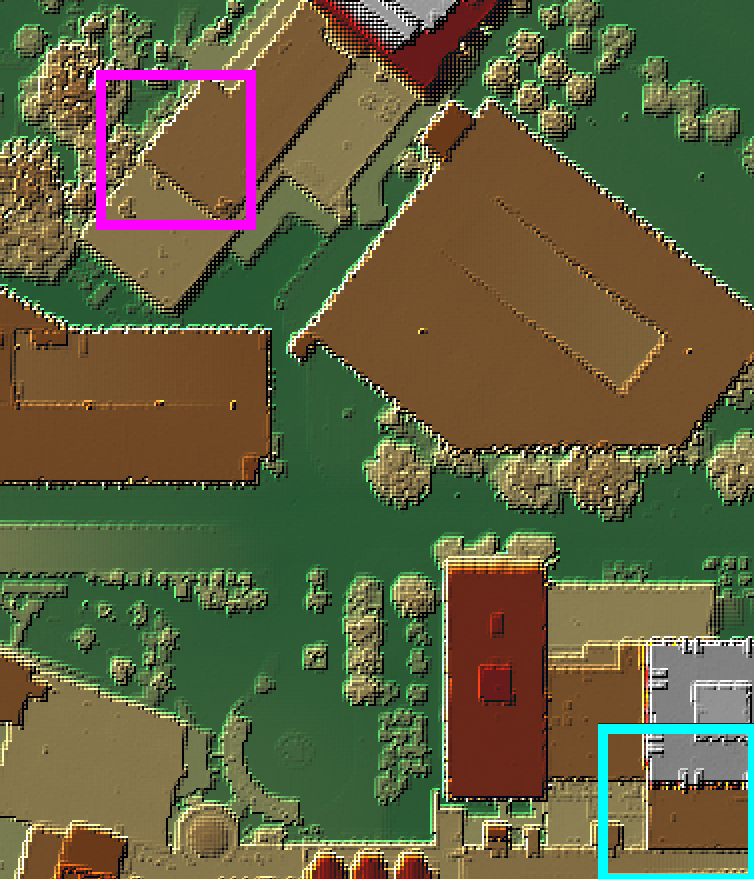}\\
        \includegraphics[width=0.24\linewidth]{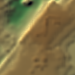}
        \includegraphics[width=0.24\linewidth]{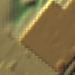}
        \includegraphics[width=0.24\linewidth]{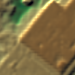}
        \textcolor{magenta}{\fboxrule=3pt\fbox{\includegraphics[width=0.24\linewidth]{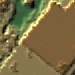}}}\\
        \includegraphics[width=0.24\linewidth]{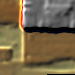}
        \includegraphics[width=0.24\linewidth]{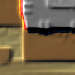}
        \includegraphics[width=0.24\linewidth]{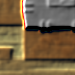}
        \textcolor{cyan}{\fboxrule=3pt\fbox{\includegraphics[width=0.24\linewidth]{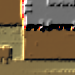}}}
    
    \vspace{-.5em}
        
        \caption{From left to right: visual results on JAX\_214 comparing SAT-NGP~\cite{billouard2024satngp}, \ourmethod, EO-NeRF~\cite{mari2023eonerf} and the ground truth.}
        \label{fig:visual}
    \end{figure}

    
    If the entire AOIs are considered, as reported in \cref{tab:exp_main} (top), \ourmethod performs slightly worse than the state of the art EO-NeRF but it is approximately $300\times$ faster. For reference, we also report all available results of other methods from the literature (EO-NeRF~\cite{mari2023eonerf}, SAT-NGP~\cite{billouard2024satngp}, Sat-Mesh~\cite{qu2023sat}, and S2P~\cite{defranchis2014automatic,facciolo2017automatic}). We see that \ourmethod is Pareto-optimal with respect to elevation MAE and training time. If instead we use available ground truth semantic maps to ignore prediction in the foliage areas, \ourmethod performance is equivalent to EO-NeRF, showing higher accuracy for structural objects, as reported in \cref{tab:exp_main} (bottom).
    We present visual results in \cref{fig:visual}, more results, as well as details on the number of Gaussians per scene and memory usage, are provided in the supplement.
    
    \subsection{Ablation and Parameter Studies}
    
    
    \begin{table}[t]
        \centering
        \setlength{\tabcolsep}{1.5pt}
        \resizebox{\linewidth}{!}{
        \begin{tabular}{cccccccccc}
        \toprule
        Shadowmap    & \xmark & \cmark & \cmark & \cmark & \cmark & \cmark & \cmark & \cmark & \cmark \\
        Sparsity     & \xmark & \xmark & \cmark & \xmark & \xmark & \xmark & \cmark & \cmark & \cmark \\
        Consistency  & \xmark & \xmark & \xmark & \cmark & \xmark & \cmark & \xmark & \cmark & \cmark \\
        Opaqueness      & \xmark & \xmark & \xmark & \xmark & \cmark & \cmark & \cmark & \xmark & \cmark \\
        \midrule
        MAE [m] $\downarrow$ & 5.03   & 1.86   & 1.83   & 1.69   & 1.79   & 1.57   & 1.76   & 1.64   & 1.54   \\
        Train [min] $\downarrow$ & 4.18  & -      & -      & -      & -      & 4.27      & -      & -      & 2.85      \\
        \bottomrule
        \end{tabular}
        }
    
    \vspace{-.5em}
    
        \caption{Ablation study of each proposed component of \ourmethod.}
        \label{tab:ablations}
    \end{table}
    
    \smallskip\noindent\textbf{Impact of the Different Losses.}
    \cref{tab:ablations} reports an ablation study of the loss terms in \ourmethod. Each column corresponds to a different ablation experiment, while each row corresponds to a different component of \ourmethod being ablated. The first row indicates whether the Shadow Mapping technique is enabled or not. The following three rows indicate, respectively, the presence of the sparsity, consistency, and opaqueness regularizers. We remark that the first column is equivalent to 3DGS with affine cameras, learnable per-camera affine color correction, and different primitives initialization. For each column, we report the grand mean elevation MAE of JAX and IARPA scenes. 
    
    To quantitatively measure the impact of each single component, we linearly regress the MAE from the presence of the components, obtaining a coefficient for each component that expresses an elevation MAE gain with respect to the base case (reported in the first column of \cref{tab:ablations}). The introduction of shadow mapping is the most impactful component, gaining $3.16$ meters of accuracy. Then, the consistency regularizer and the opaqueness regularizer further improve the accuracy of \ourmethod by $0.20$ and $0.09$ meters, respectively. Lastly, the sparsity regularizer, while being necessary for achieving efficient training as shown in the last row of \cref{tab:ablations}, also reports an improvement of $0.04$ meters. Hence, all components independently contribute to the quality of the geometry reconstruction.
    
    \begin{figure}
        \centering
        \includegraphics[width=0.9\linewidth]{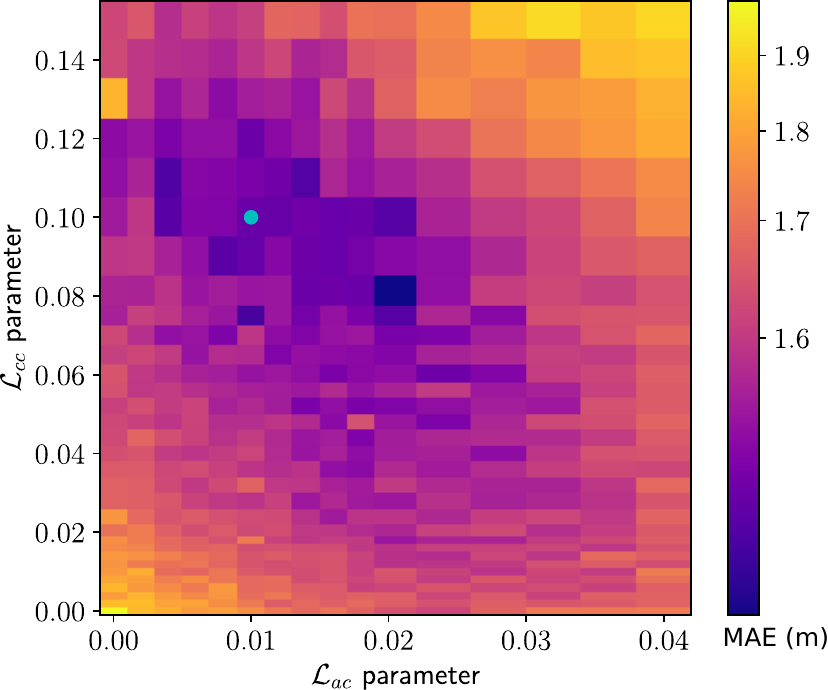}
        
        \vspace{-0.5em}

        \caption{View consistency regularization parameter ablation study. Selected parameter set is shown with the cyan dot. Estimation performed on the JAX\_260 sequence.}
        \label{fig:ablation_reg}
    \end{figure}
    
    \smallskip\noindent\textbf{Regularization Parameters.}
    \cref{fig:ablation_reg} shows the results of a grid search on the coefficients of $\mathcal{L}_{cc}$ and $\mathcal{L}_{ac}$ in \cref{eq:regularizedproblem}. It shows that both  altitude regularization and  color regularization are necessary to achieve the best performance. We remark that, in order to reduce overfitting to a particular scene, we choose the same \virg{round coefficients} for all scenes. 
    
\begin{figure}
        \centering
        \includegraphics[width=\linewidth]{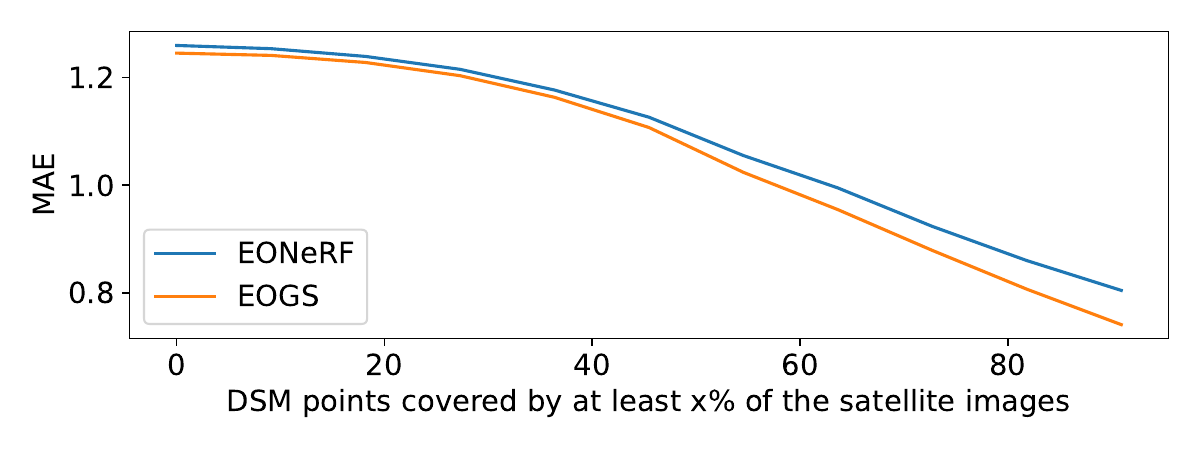}
        
        \vspace{-1em}
        
        \caption{Impact of the visibility on the performance (using foliage mask).}
        \label{fig:ablation_vis}
    \end{figure}

    \smallskip\noindent\textbf{Impact of visibility.}
    \cref{fig:ablation_vis} shows the impact of the visibility (\ie the number of cameras that can see a given point of the scene) on the performance. While \ourmethod and EO-NeRF are comparable on average, this test shows that \ourmethod performs better for regions that are visible in most images but struggles in the regions observed in only a few images.
    
    \section{Conclusion}
    
    This study presents EOGS, the first Gaussian-Spatting-based framework for earth observation. By introducing remote sensing requirements in 3DGS, such as shadow modeling and camera-specific corrections, EOGS achieves comparable accuracy to the existing state-of-the-art method, EO-NeRF, while being over 300 times faster. Hence, EOGS will be a practical solution for near-future large-scale datasets.
    
    Our analysis shows that EOGS excels in high-coverage regions, where it produces fine details at a fraction of the computational cost. Addressing regions with low image coverage or with foliage, by adding additional regularization or employing better initialization schemes, could further improve EOGS accuracy.
    
    
    \subsubsection*{Acknowledgements}
    This publication is part of the project PNRR-NGEU which has received funding from the MUR - DM 352/2022. This work was partially supported by the European Union under the Italian National Recovery and Resilience Plan (NRRP) of NextGenerationEU, partnership on ``Telecommunications of the Future'' (PE00000001 - program ``RESTART''). It was performed using HPC resources from GENCI-IDRIS (grant 2024-AD011012453R3). Centre Borelli is also with Universit\'{e} Paris Cit\'{e}, SSA and INSERM.

{
    \small
    \bibliographystyle{ieeenat_fullname}
    \bibliography{main}
}
\end{document}